\documentclass[10pt,twocolumn,letterpaper]{article}

\usepackage{iccv}

\usepackage{times}
\usepackage{epsfig}
\usepackage{graphicx}
\usepackage{amsmath}
\usepackage{amssymb}

\makeatletter
\@namedef{ver@everyshi.sty}{}
\makeatother
\usepackage{tikz}

\usepackage{booktabs}
\usepackage{comment}
\usepackage{color}
\usepackage{array}
\usepackage{multirow}

\usepackage[pagebackref=true,breaklinks=true,letterpaper=true,colorlinks,bookmarks=false]{hyperref}

 \iccvfinalcopy 

\ificcvfinal\fi

\begin{document}

\title{A Vanilla Multi-Task Framework for Dense Visual Prediction\\ Solution to 1$^{st}$ VCL Challenge -- Multi-Task Robustness Track}

\author{Zehui Chen$^1$ \quad Qiuchen Wang$^1$ \quad Zhenyu Li$^{2}$ \quad Jiaming Liu$^{3}$ \quad Shanghang Zhang$^{3}$ \quad Feng Zhao$^{1}$\\
$^{1}$University of Science and Technology of China \\
$^{2}$King Abdullah University of Science and Technology \\
$^{3}$Peking University \\
{\small\texttt{lovesnow@mail.ustc.edu.cn \hspace{10pt} fzhao956@ustc.edu.cn}}
}

\maketitle
\ificcvfinal\thispagestyle{empty}\fi

\begin{abstract}
	In this report, we present our solution to the multi-task robustness track of the 1$^{st}$ visual continual learning (VCL) challenge at ICCV 2023 Workshop. We propose a vanilla framework named UniNet that seamlessly combines various visual perception algorithms into a multi-task model. Specifically, we choose DETR3D, Mask2Former, and BinsFormer for 3D object detection, instance segmentation, and depth estimation tasks, respectively. The final submission is a single model with InternImage-L backbone, and achieves 49.6 overall score (29.5 Det mAP, 80.3 mTPS, 46.4 Seg mAP, and 7.93 silog) on SHIFT validation set. Besides, we provide some interesting observations in our experiments which may facilitate the development of multi-task learning in dense visual prediction. 
\end{abstract}

\section{Introduction}

We introduce a vanilla multi-task dense visual prediction framework for 1$^{st}$ visual continual learning (VCL) Challenge -- multi-task robustness track at ICCV 2023 Workshop. It introduces SHIFT dataset \cite{sun2022shift}, a multi-task driving dataset featuring the most important perception tasks under a variety of conditions and with a comprehensive sensor setup. It consists of 4,850 video sequences (2.5M image frames), as well as rich annotations, covering 2D/3D detection and tracking, depth estimation, instance/semantic segmentation tasks. The multi-task robustness track evaluates the overall model performance on the 3D detection, instance segmentation, and depth estimation tasks. 
\section{Method}

\begin{figure}[!t]
\begin{center}
\includegraphics[width=0.45\textwidth]{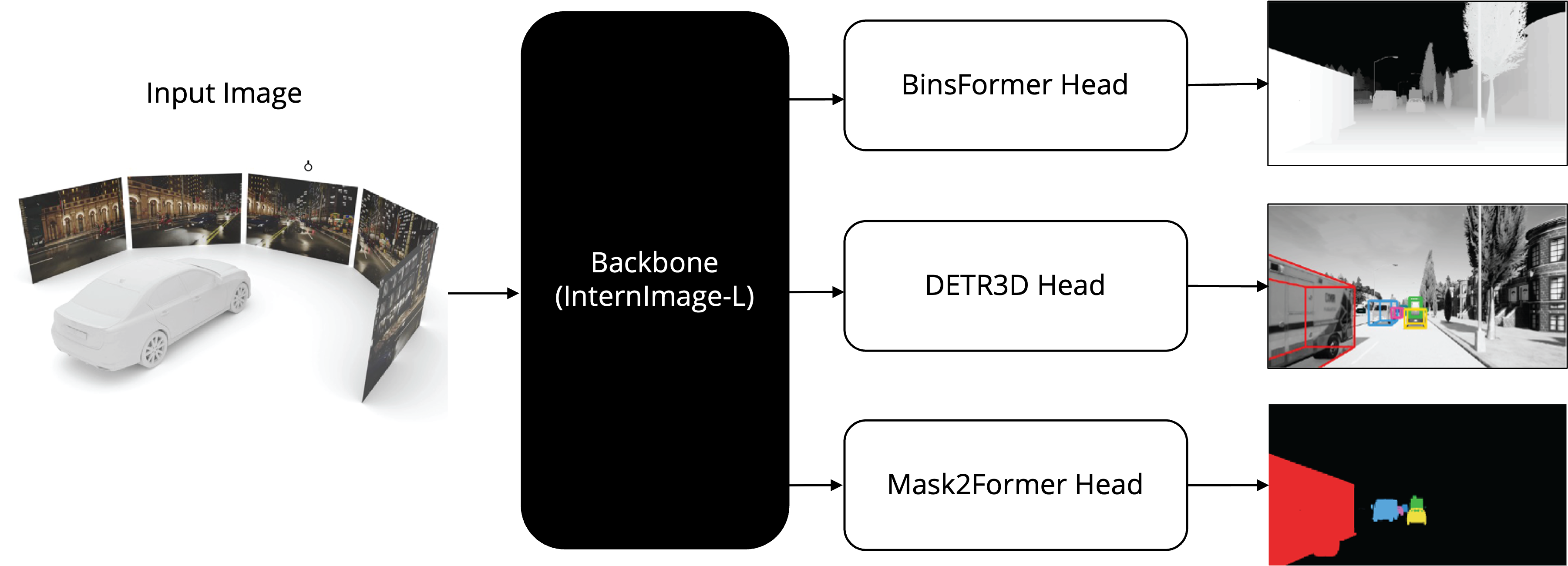}
\end{center}
\caption{Framework Overview of UniNet, which consists of an image backbone and 3 respective task heads.}
\label{fig:framework}
\end{figure}

In this section, we first describe each component of our approach and then introduce how we simply combine them to form UniNet.

\subsection{Instance Segmentation}

Instance segmentation aims to detect the interested objects and provide the pixel-level masks \cite{he2017mask,kirillov2020pointrend,wei2023stronger,chen2022deliberated}. We follow the original implementation of Mask2Former \cite{cheng2022masked}. Thanks to the separate training schedule, we are able to perform strong data augmentations on Mask2Former, such large scale image resize and crop.

\subsection{Depth Estimation}

Monocular depth estimation stands an difficult and critical task in autonomous driving \cite{li2022depthformer,bhat2021adabins}. We follow the official implementation of BinsFormer \cite{li2022binsformer}, state-of-the-art approach on KITTI depth estimation benchmark. It shares a similar architecture of Mask2Former. However, we do not shared much neural parameters in our final model design for simplicity. 

\subsection{3D Object Detection}

Since the dataset only provides the front view of the images, this task is actually a monocular 3D detection task, rather than multi-view one. Thanks to the generalization ability of DETR3D \cite{wang2022detr3d}, we are able to adapt it to single view setting easily. Considering that the detection range is extremely far (120m vs 50m, more than 2 $\times$ larger) compared to standard nuScenes dataset \cite{caesar2020nuscenes}, we select the P2-P5 levels rather than P3-P6 levels in original DETR3D implementation from FPN to preserve the fine-grained feature from the backbone feature extractor. Besides, the original DETR3D is initialized from the FCOS3D pretrained checkpoint on nuScenes dataset. Such a strategy is to ensure that the image backbone is able to extract depth-related features, which avoids the model collapes during training. In the spirit of this, we utilizes the BinsFormer checkpoint trained on the SHIFT depth estimation track to initialize the DETR3D image backbone. We also tried FCOS3D \cite{wang2021fcos3d} on this task, however, the final performance is much lower than DETR3D (10 mAP vs 30 mAP, perhaps there is some missing details for FCOS3D).

Due to the time limitation, we do not tried other approaches \cite{chen2023ddod,huang2022bevdet4d,li2022bevformer}, but we believe more advanced techniques, such as long-time sequence fusion \cite{park2022time,han2023exploring}, stronger feature aggregation module \cite{lin2022sparse4d,chen2022graph}, and knowledge distillation \cite{chen2022bevdistill,chang2022detrdistill} will definitely improve the final detection performance. We leave it for future work.

\subsection{UniNet}

In this section, we introduce UniNet, a vanilla multi-task learning framework for visual prediction. It consists of a shared image backbone and 3 separate heads, specialized for each task. An overview of our framework is shown in Figure \ref{fig:framework}. The training schedule can be divided into two steps: 
\begin{itemize}
	\item Train the above three models separatedly and then merge them into one single model.
	\item Fine-tune the model on three tasks concurrently for a few epochs to adapt the merge setting.
\end{itemize} 
In terms of the model merge, we simply average the weight of image backbone from respective detectors. Such a training schedule enables us to enjoy the rich data augmentation strategy designed for each task, meanwhile achieves the goal of a multi-task learning framework.
\section{Experiments}

\subsection{Implementation Details}

We implement our methods based on the official repository MMDetection3D, MMDetection, Depth-ToolBox. Thanks to the extendability of MM framework, we can easily integrate Mask2Former, BinsFormer into the MMDetection3D. The detection range is limited to [0m, 120m], [-60m, 60m], and [-3m, 3m] for x, y, z-axis during training and testing process. If not specified, we follow the default training setting for each model. As for the finetune step, we use the AdamW optimizer with Cosine annealing learning rate schedule. The model is trained at a learning rate of 4e-5 for 12 epochs. Due to the time limitation, we set the training interval to 2 for this stage, which means we only finetune the model on the 1/2 subset of the training set. We also implement test-time augmentation during testing for instance segmentation and depth estimation tasks, following the implementation in \cite{huang20201st}.

\subsection{Main Results}

\begin{figure}[!h]
\begin{center}
\includegraphics[width=0.5\textwidth]{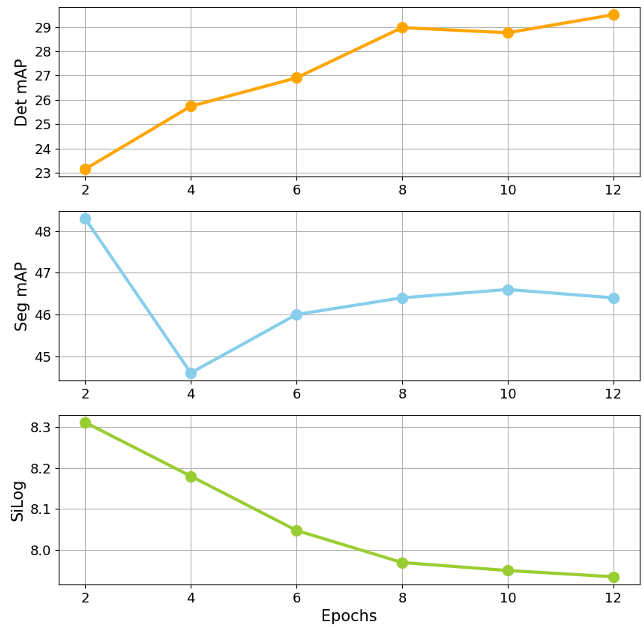}
\end{center}
\caption{The per-task evaluation of each task during finetuning process of UniNet, including 3D object detection, instance segmentation, and depth estimation.}
\label{fig:ft_res}
\end{figure}

In this section, we report the model performance on the SHIFT validation subset for each individual models and the final finetuned model in Table \ref{tab:main_result}. It can be seen that our UniNet is able to surpass all its individual models by a large margin, which validate the necessity of the joint training paradigm for multi-task learning. 

\begin{table}[!t]
	\caption{Overall performance of our final UniNet model and each individual component before merging on SHIFT validation subset.}
	\label{tab:main_result}
	\vskip 0.5em
	\centering
	\setlength{\tabcolsep}{2.0mm}
		\begin{tabular}{c |c c | c |c }
			\toprule
			Model & Det mAP & mTPS & Seg mAP & SiLog \\
			\midrule
			DETR3D & 23.9 & 76.7 & - & - \\
			Mask2Former & - & - & 39.5 & -\\
			BinsFormer & - & - & - & 8.83\\
			\midrule
			UniNet & \textbf{29.5} & \textbf{80.4} & \textbf{46.4} & \textbf{7.93}\\
			\bottomrule
		\end{tabular}
\end{table}

\begin{figure*}[!h]
\begin{center}
\includegraphics[width=1.0\textwidth]{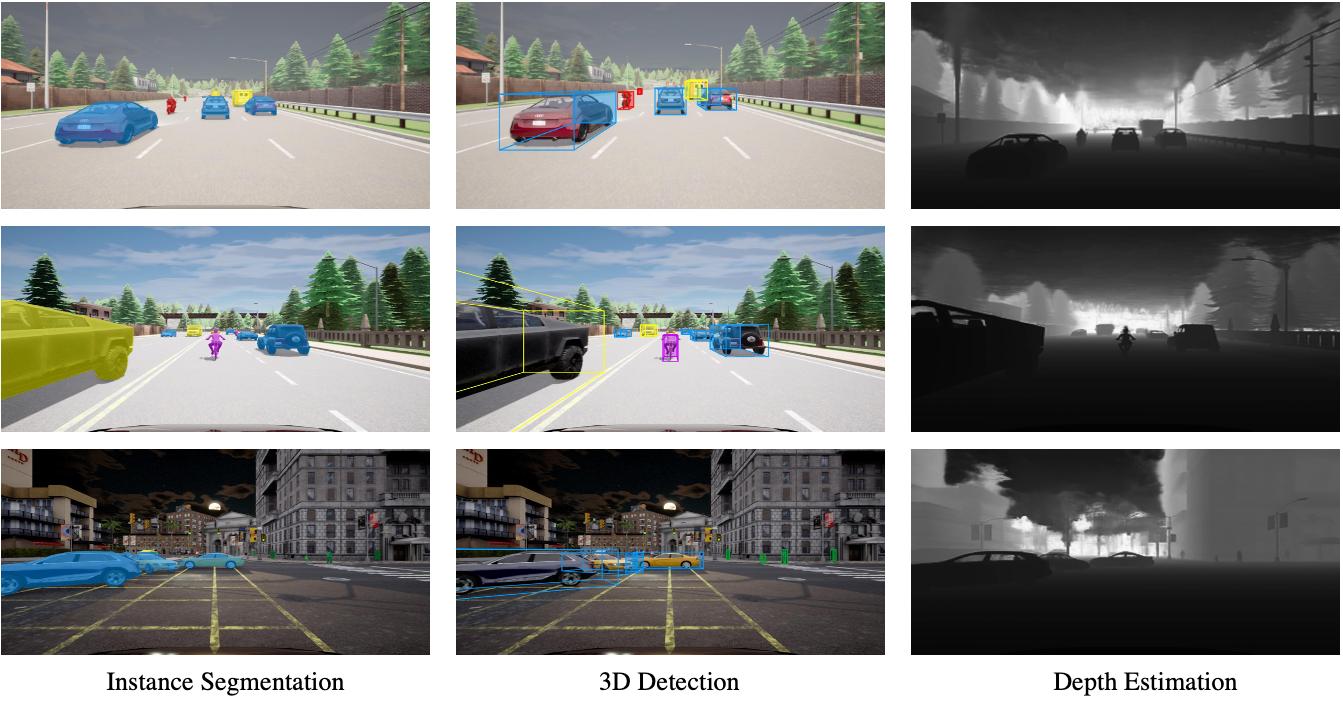}
\end{center}
\caption{The visualization of UniNet prediction by our final model with InternImage-L backbone on instance segmentation, 3D detection, and depth estimation tasks.}
\label{fig:vis}
\end{figure*}

\subsection{Discussion}

In addition to the overall performance of our model, we also provide a detailed performance evaluation during the finetuning step in Figure \ref{fig:ft_res}. From the figure, we can see that our model gets better performance at both 3D detection and depth estimation tasks. However, for the instance segmentation, the model get 48.3 mAP at the 2rd epoch, and then incurs a drastic drop, more than 3 mAP. Despite the final performance converges to 46.3 mAP, it is still lower than the initial performance. We attribute the reason to the weaker data augmentation during the finetuning process. A possible solution can be lower the learning rate in the instance segmentation head, but we have no time to validate it. Besides, such a strategy still does not resolve the problem directly. We expect more elegant approaches to circumvent this issue, which lies an important barrier for multi-task learning. 

\subsection{Visualization}

We also visualize some predictions by UniNet on SHIFT mini validation subset in Figure \ref{fig:vis}.

\section{Conclusion}

In this report, we present a competitive multi-task framework for dense visual prediction and achieves xx overall score on the multi-task robustness track of the VCL challenge at ICCV 2023 Workshop. We hope our work could inspire more researches on the large scale SHIFT dataset in the future.


{\small
\bibliographystyle{ieee_fullname}
\bibliography{egbib}
}

\end{document}